\documentclass[letterpaper, 10 pt, conference]{ieeeconf}
\IEEEoverridecommandlockouts
\usepackage{cite}
\usepackage{amsmath,amssymb,amsfonts}
\usepackage{algorithmic}
\usepackage{algorithm}
\usepackage{array}
\usepackage[caption=false,font=normalsize,labelfont=sf,textfont=sf]{subfig}
\usepackage{textcomp}
\usepackage{stfloats}
\usepackage{url}
\usepackage{verbatim}
\usepackage{fancyvrb}
\usepackage{graphicx}
\usepackage{booktabs}
\usepackage{multirow}
\usepackage{placeins}
\usepackage{xcolor}
\usepackage{latexml}
\PassOptionsToPackage{hyphens}{url}
\usepackage{xurl}
\usepackage[breaklinks=true,hidelinks]{hyperref}
\hypersetup{
  pdfauthor={Junhui Wang, Wei Yang, Xinyao Li, Ningjing Fan, Yuehao Yin, Xuecheng Chen, Chao Gao},
  pdftitle={},
  pdfsubject={},
  pdfkeywords={}
}
\usepackage{titletoc}

\DefineVerbatimEnvironment{NavTree}{Verbatim}{%
  fontsize=\scriptsize,
  frame=single,
  framesep=2mm,
  framerule=0.2pt
}

\def\BibTeX{{\rm B\kern-.05em{\sc i\kern-.025em b}\kern-.08em
    T\kern-.1667em\lower.7ex\hbox{E}\kern-.125emX}}
\newcommand{\partialsupport}{\ensuremath{\triangle}}


\title{\LARGE \bf NavArena: Automated Construction of Goal-Oriented Navigation Benchmarks from 3D Gaussian Splatting Reconstructions}

\iflatexml
  \author{Junhui Wang \and Wei Yang \and Xinyao Li \and Ningjing Fan \and
  Yuehao Yin \and Xuecheng Chen \and
  Chao Gao\textsuperscript{*}\thanks{\textsuperscript{*}Corresponding author.}}
\else
  \author{%
\author{Junhui Wang, Wei Yang, Xinyao Li, Ningjing Fan,\\
Yuehao Yin, Xuecheng Chen, and Chao Gao\textsuperscript{*}%
\thanks{\textsuperscript{*}Corresponding author.}}
  \authorblockN{Junhui Wang, Wei Yang, Xinyao Li, Ningjing Fan,\\
  Yuehao Yin, Xuecheng Chen, and Chao Gao\textsuperscript{*}}%
  \thanks{\textsuperscript{*}Corresponding author.}}
\fi

\begin{document}
\maketitle
\begin{abstract}
Fixed 3D Gaussian Splatting (3DGS) reconstructions provide realistic novel views but lack the traversability constraints, valid goals, and closed-loop protocols required for navigation evaluation. We introduce NavArena, an automated framework that transforms fixed 3DGS reconstructions into benchmarks for goal-oriented visual navigation. NavArena integrates a frozen 3DGS model for egocentric RGB-D rendering, an occupancy costmap derived from Gaussian density and height statistics for reachability and collision queries, and semantic goal candidates lifted from multi-view open-vocabulary masks. These components support the automatic generation and unified closed-loop evaluation of goal-oriented navigation episodes. Across more than 2{,}000 scenes, NavArena generates 22.2 million expert trajectories. Spatial and semantic evaluations assess the derived navigation representations, while policy rollouts demonstrate the diagnostic value of the unified evaluation protocol. NavArena enables scalable and reproducible navigation evaluation on large-scale 3DGS reconstructions, and all benchmark-generation tools, evaluation protocols, and derived assets will be released publicly.
\end{abstract}


\section{Introduction}

Goal-oriented visual navigation requires closed-loop evaluation: an agent acts from observations, receives new views as it moves, and must reach a metric, visual, or semantic target without violating motion constraints~\cite{batra2020objectnav}. A useful benchmark must therefore provide not only realistic observations, but also traversability, collision feedback, reachable goals, and consistent success criteria throughout a rollout.

Large-scale navigation benchmarks typically obtain these capabilities from curated simulators, meshes, and simulator-native annotations~\cite{savva2019habitat,kolve2017ai2thor,xia2018gibson,makoviychuk2021isaac}. Meanwhile, 3D Gaussian Splatting (3DGS) has enabled high-quality view synthesis at practical rendering rates~\cite{kerbl2023gaussian}, and large collections of 3DGS reconstructions are becoming available~\cite{sage3d2026,ma2025scenesplatpp}. These reconstructions offer a scalable source of visually realistic scenes, but appearance alone does not define a navigation environment. A 3DGS reconstruction provides neither explicit free space and collision constraints nor verified metric, visual, and semantic goals. Existing pipelines can recover some of these signals through mesh reconstruction, NavMesh construction, or scene-specific semantic optimization, but they do not directly provide an automated benchmark-construction procedure for heterogeneous 3DGS collections~\cite{xia2026habitatgs}.

This gap raises a practical question: can a fixed 3DGS reconstruction be augmented with the spatial and semantic interfaces required for reproducible closed-loop navigation evaluation, without first converting it into a complete physical simulator? Addressing this question requires three capabilities in a common coordinate system: egocentric rendering, collision-aware motion constraints, and reachable task-specific goals. The resulting interface should also support automatic episode generation and apply consistent actions, termination conditions, and metrics across navigation tasks.

We introduce NavArena, a framework that automatically constructs goal-oriented navigation benchmarks from fixed 3DGS reconstructions. NavArena preserves the frozen 3DGS representation for egocentric RGB-D rendering and derives two aligned navigation layers. First, an occupancy costmap estimated from Gaussian density and height statistics supports reachability analysis, trajectory generation, and collision queries. Second, a semantic layer lifts multi-view open-vocabulary masks into Object Navigation (ObjectNav) goal candidates without scene-specific training or semantic-field optimization~\cite{kerr2023lerf}. Together, the visual, spatial, and semantic layers provide a unified interface for Point Navigation (PointNav), Image Navigation (ImageNav), and ObjectNav. NavArena avoids per-scene mesh repair~\cite{guedon2024sugar} and manual semantic annotation, while deliberately focusing on navigation benchmark interfaces rather than rigid-body dynamics or general robot physics.

The main contributions are summarized as follows.

\begin{itemize}
  \item We propose an automated method that augments fixed 3DGS reconstructions with traversability, collision queries, and semantic goal candidates, without per-scene mesh repair, semantic annotation, or semantic-field optimization.

  \item We develop a unified benchmark-construction and closed-loop evaluation pipeline for PointNav, ImageNav, and ObjectNav, including task-specific episode generation under a common action and metric interface.

  \item We demonstrate scalability across more than 2{,}000 heterogeneous scenes, generating 22.2 million expert trajectories and supporting closed-loop policy evaluation.
\end{itemize}

\begin{table*}[t]
  \caption{%
    Comparison of navigation-environment construction pipelines based on the capabilities reported by their respective sources.
    NavArena focuses on automated benchmark construction from fixed 3DGS assets.
  }
  \label{tab:comparison}
  \centering
  \small
  \setlength{\tabcolsep}{1.4pt}
  \begin{tabular}{@{}>{\raggedright\arraybackslash}m{0.19\textwidth}
                  >{\centering\arraybackslash}m{0.080\textwidth}
                  >{\centering\arraybackslash}m{0.08\textwidth}
                  >{\centering\arraybackslash}m{0.05\textwidth}
                  >{\centering\arraybackslash}m{0.110\textwidth}
                  >{\centering\arraybackslash}m{0.110\textwidth}
                  >{\centering\arraybackslash}m{0.110\textwidth}
                  >{\centering\arraybackslash}m{0.110\textwidth}@{}}
    \toprule
    Platform & \shortstack{Rendering\\model} & \shortstack{Collision\\model} & Scale & \shortstack{Automated\\onboarding} & \shortstack{Automated\\goal extraction} & \shortstack{Goal-nav\\expert traj.} & \shortstack{Closed-\\loop evaluation} \\
    \midrule
    Habitat\,+\,HM3D~\cite{savva2019habitat,ramakrishnan2021hm3d} & Mesh & NavMesh & 1K & \(\times\) & \(\times\) & \checkmark & \checkmark \\
    Habitat-GS~\cite{xia2026habitatgs} & 3DGS & NavMesh & -- & \partialsupport & \(\times\) & \checkmark & \checkmark \\
    GS-Playground~\cite{Jia2026_EHVJRIBC} & 3DGS & Mesh & -- & \(\partialsupport\) & \(\times\) & \(\times\) & \(\times\) \\
    DISCOVERSE~\cite{jia2025discoverse} & 3DGS & Mesh & -- & \(\partialsupport\) & \(\times\) & \(\times\) & \(\times\) \\
    GaussGym~\cite{escontrela2025gaussgym} & 3DGS & Mesh & -- & \(\partialsupport\) & \(\times\) & \(\times\) & \(\times\) \\
    \midrule
    \textbf{NavArena} & \textbf{3DGS} & \textbf{Costmap} & \textbf{2K+} & \textbf{\checkmark} & \textbf{\checkmark} & \textbf{\checkmark} & \textbf{\checkmark} \\
    \bottomrule
  \end{tabular}
  \vspace{2pt}

  \begin{minipage}{0.9\textwidth}
    \scriptsize\emph{Notes:} Scale follows each source's reporting convention and is therefore not directly comparable across rows. Automated scene onboarding denotes benchmark construction from a valid, metric-calibrated scene asset, without manual scene-specific mesh repair, semantic labeling, or per-scene semantic optimization. Goal-nav expert traj. denotes target-conditioned trajectory generation. \checkmark\ denotes reported support, \partialsupport\ denotes setup-dependent or partial support, and \(\times\) denotes support not reported in the cited source.
  \end{minipage}
\end{table*}

\section{Related Work}
\label{sec:related}

\subsection{Goal-Oriented Navigation Benchmarks and Simulators}
\label{sec:related:navigation}
\label{sec:related:platforms}

Goal-oriented navigation requires an agent to reach a metric location, reference view, or semantic object category through closed-loop interaction~\cite{batra2020objectnav}. Despite their different goal specifications, they rely on the same evaluation infrastructure: egocentric observations, collision-aware transitions, reachable goals, and consistent termination criteria. Existing platforms provide this infrastructure through interactive environments~\cite{kolve2017ai2thor}, reconstructed indoor scenes~\cite{xia2018gibson,savva2019habitat,ramakrishnan2021hm3d}, procedurally generated assets~\cite{deitke2022procthor}, or GPU-based physics simulation~\cite{makoviychuk2021isaac}. These platforms enable controlled evaluation, but their construction pipelines typically presuppose curated geometry, simulator-native annotations, or scene-specific asset preparation. NavArena addresses a different setting: it derives the interfaces required for navigation directly from fixed 3DGS reconstructions that were not created as simulator-ready assets.

\subsection{3DGS-Based Embodied Environments}
\label{sec:related:neural}

3D Gaussian Splatting provides high-quality novel-view synthesis at practical rendering rates~\cite{kerbl2023gaussian}, making it a useful visual representation for embodied simulation and navigation. Existing studies have integrated Gaussian rendering into navigation simulators~\cite{xia2026habitatgs}, used reconstructed scenes for visual navigation and policy reasoning~\cite{lei2025gaussnav}, and coupled Gaussian appearance models with robot-learning or physics backends~\cite{jia2025discoverse,escontrela2025gaussgym}. Together, these efforts demonstrate the value of 3DGS for observation synthesis, navigation, and robot learning. Their primary focus, however, is rendering integration, policy development, or physics-enabled simulation rather than a general procedure for converting fixed 3DGS assets into navigation-ready benchmarks.

Fixed 3DGS reconstructions do not explicitly encode traversability, collision-valid transitions, or reachable task goals. NavArena derives these interfaces from the frozen asset through an occupancy costmap and aligned semantic candidates, which support both episode generation and closed-loop evaluation. Table~\ref{tab:comparison} therefore compares environment-construction interfaces; NavArena targets automated goal-navigation benchmark construction rather than general-purpose physics simulation.

\subsection{Semantic Grounding for Object Navigation}
\label{sec:related:semantic}

ObjectNav requires semantic evidence to be associated with discrete instances and reachable, collision-valid terminal states. Existing approaches distill visual-language features into Gaussian representations~\cite{li2025scenesplat,ma2025scenesplatpp}, optimize scene-specific semantic representations~\cite{qin2024langsplat,li2025instancegaussian}, or lift multi-view segmentation evidence into an existing reconstruction~\cite{shen2024flashsplat}. NavArena builds on the lifting paradigm to avoid scene-specific semantic training. It further aggregates the fused evidence into instance candidates and retains only candidates with reachable, collision-valid goal states. The resulting semantic layer is therefore designed for reproducible ObjectNav episode construction rather than dense scene understanding alone.

\section{Unified Scene Representation}
\label{sec:overview}

\begin{figure*}[t]
  \centering
  \includegraphics[width=\textwidth]{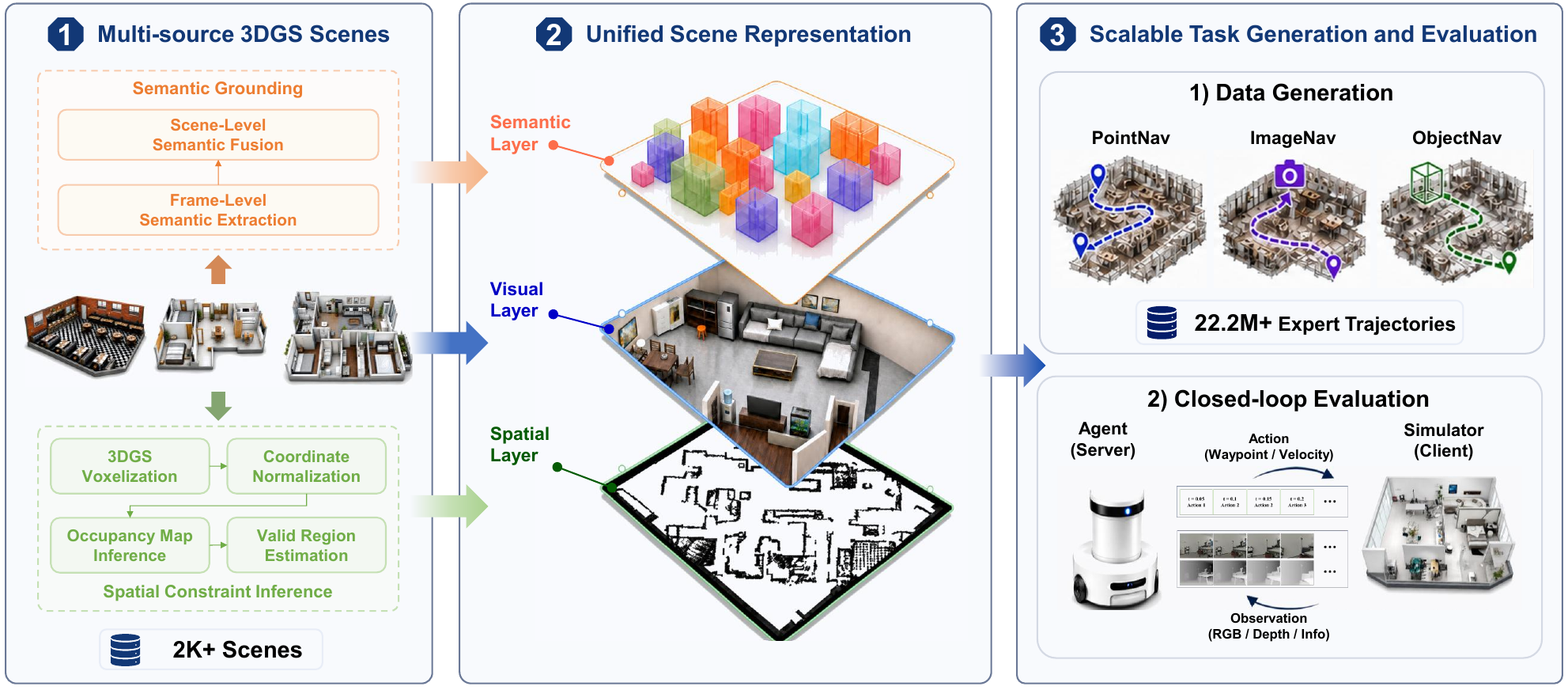}
  \caption{NavArena construction pipeline.
    A frozen 3DGS model provides egocentric rendering, while separately inferred spatial and semantic layers define where the agent can move and which goals are valid.
    PointNav, ImageNav, and ObjectNav episodes are accepted only after reachability, collision, and goal-validity checks, then evaluated through the same closed-loop protocol.
  }
  \label{fig:overview}
\end{figure*}

Figure~\ref{fig:overview} provides an overview of the NavArena pipeline, from multi-source 3DGS scene conversion to navigation-ready scene representations, scalable task generation, and closed-loop evaluation.

NavArena treats each converted reconstruction as a benchmark asset rather than a full physical simulator. For each retained scene, the asset is defined as
\[
    \mathcal{S}=(\mathcal{G}, \mathcal{M}_{\mathrm{occ}}, \mathcal{I}),
\]
where \(\mathcal{G}\) is the renderable 3DGS model, \(\mathcal{M}_{\mathrm{occ}}\) is the navigation constraint layer, and \(\mathcal{I}\) is the set of semantic object-instance candidates.

In a shared metric frame, \(\mathcal{G}\) renders egocentric observations, \(\mathcal{M}_{\mathrm{occ}}\) supports navigation-validity queries, and \(\mathcal{I}\) provides labeled 3D instance centers. This separation avoids requiring watertight geometry, rigid-body physics, or simulator-native semantics.

A benchmark instance is \(\mathcal{B}=(\mathcal{S},\mathcal{D},\Pi)\), where \(\mathcal{D}\) contains episodes, trajectories, and goals, and \(\Pi\) specifies the action interface, termination rules, metrics, and closed-loop protocol. The latter two components remain task-specific.

\section{Navigation-Ready 3DGS Scene Construction}
\label{sec:scenes}

\subsection{Overview}
\label{sec:scenes:rep_design}

Given a frozen 3D Gaussian Splatting model \(\mathcal{G}=\{g_i\}_{i=1}^{N_{\mathrm{G}}}\), camera poses, and an open-vocabulary ObjectNav category set, NavArena constructs \(\mathcal{S}\) through coordinate normalization, height-aware occupancy inference, valid-region estimation, and semantic grounding. We assume that the input reconstruction is valid and that the 3DGS model and camera poses share a consistent metric coordinate frame.

The resulting asset exposes two primitive queries to the benchmark protocol. At time step \(t\), let \(q_t\) denote the agent pose. Rendering returns the RGB-D observation
\[
    o_t=(I_t,D_t)=\mathcal{R}(\mathcal{G},q_t)
\]
where \(I_t\) is the rendered RGB image and \(D_t\) is the pixel-aligned metric depth map. The transition model computes
\[
    q_{t+1} = \mathcal{T}(q_t, u_t, \mathcal{M}_{\mathrm{occ}}).
\]
Here, \(u_t\) denotes the control input. The transition query evaluates collision and motion validity through \(\mathcal{M}_{\mathrm{occ}}\), while \(\mathcal{G}\) remains responsible only for observation rendering.

\subsection{Spatial Constraint Inference}
\label{sec:scenes:constraints}

Spatial-layer construction voxelizes Gaussian opacity, normalizes the coordinate frame, and derives an agent-height-aware costmap and valid sampling region.

\paragraph{3DGS Voxelization}
Voxelization converts the frozen reconstruction into a coarse occupied volume. The geometric pass treats the 3DGS as an opacity field and ignores color and spherical harmonic attributes. For each Gaussian \(g_i=(\boldsymbol{\mu}_i,\boldsymbol{\eta}_i,\mathbf{s}_i,a_i,\cdots)\), \(\boldsymbol{\mu}_i\) is its center, \(\boldsymbol{\eta}_i\) is its rotation quaternion, \(\mathbf{s}_i\) contains its log-scales, and \(a_i\) is its opacity logit. NavArena recovers physical scales and opacity as \(\boldsymbol{\sigma}_i=\exp(\mathbf{s}_i)\) and \(\alpha_i=(1+\exp(-a_i))^{-1}\). Each overlapping Gaussian contributes to a voxel \(\nu\), centered at \(\mathbf{x}_{\nu}\), as
\begin{equation}
    \label{eq:gaussian-voxel-contribution}
    \chi_{i,\nu}^{2}=\sum_{k\in\{x,y,z\}}\frac{d_{i,\nu,k}^{2}}{\sigma_{i,k}^{2}},
    \qquad
    \rho_{i,\nu}=\alpha_i\exp\left(-\frac{1}{2}\chi_{i,\nu}^{2}\right),
\end{equation}
In~\eqref{eq:gaussian-voxel-contribution}, \(\chi_{i,\nu}^{2}\) is the squared Mahalanobis distance, \(\mathbf{d}_{i,\nu}=\mathbf{R}_i^{\top}(\mathbf{x}_{\nu}-\boldsymbol{\mu}_i)\), and \(\mathbf{R}_i\) is induced by \(\boldsymbol{\eta}_i\). Let \(\mathcal{N}(\nu)\) be the set of Gaussians whose truncated support overlaps \(\nu\). Treating the bounded contributions \(\rho_{i,\nu}\in[0,1]\) as independent occupancy evidence, the effective voxel opacity \(A_{\nu}\) and binary occupancy \(V(\nu)\) are defined as
\begin{equation}
    \label{eq:voxel-occupancy}
    A_{\nu}=1-\prod_{i\in\mathcal{N}(\nu)}\left(1-\rho_{i,\nu}\right),
    \qquad
    V(\nu)=\mathbf{1}[A_{\nu}\geq\tau_{\mathrm{occ}}],
\end{equation}
where \(\tau_{\mathrm{occ}}\) denotes the opacity threshold. The voxel field in~\eqref{eq:voxel-occupancy} provides the evidence used to derive the navigation costmap.

\paragraph{Coordinate Normalization}
Given the assumed input frame, coordinate normalization places heterogeneous reconstructions in a common navigation frame using only a rigid transform; it does not estimate or calibrate physical scale. NavArena estimates the dominant support plane from occupied voxels using RANSAC, aligns its normal with the gravity axis, and translates the plane to \(z=0\). The same rigid transform is applied to the voxel field, and 3DGS scene, producing a normalized voxel field \(V^{\star}\) paired with a canonically aligned 3DGS.

\paragraph{Occupancy Map Inference}
Occupancy map inference compresses the aligned volume into the 2D structure used by online navigation. Let \(h_{\min}\) and \(h_{\max}\) denote the lower and upper vertical bounds of the collision body relative to the aligned ground plane. The embodiment-specific body interval is \(\mathcal{H}_{\mathrm{body}}=[h_{\min},h_{\max}]\); its bounds are supplied by the robot configuration rather than fixed by NavArena. Let \(\mathcal{Z}_{i_z}\) denote the vertical extent of voxel index \(i_z\). The raw planar occupancy map marks a cell as occupied when a voxel in its vertical column intersects the configured body interval:
\begin{equation}
    \label{eq:raw-planar-occupancy}
    M_{\mathrm{raw}}(i_x,i_y)=
    \bigvee_{i_z:\,\mathcal{Z}_{i_z}\cap\mathcal{H}_{\mathrm{body}}\neq\varnothing}
    V^{\star}(i_x,i_y,i_z).
\end{equation}
Equation~\eqref{eq:raw-planar-occupancy} adapts to each embodiment through \(\mathcal{H}_{\mathrm{body}}\).

\paragraph{Valid Region Estimation}
Valid region estimation restricts navigation to regions for which the reconstruction provides spatial evidence. DBSCAN first removes isolated occupied-voxel clusters. Let \(\mathcal{X}_{\mathrm{proj}}\subset\mathbb{R}^{2}\) be the ground-plane projection of the remaining occupied voxels. An \(\alpha\)-shape concave hull defines the reconstruction-supported domain and its binary support mask:
\begin{equation}
\label{eq:support-mask}
\begin{aligned}
    \Omega_{\mathrm{sup}}
    &=\operatorname{AlphaShape}\!\left(
      \mathcal{X}_{\mathrm{proj}};\alpha_{\mathrm{hull}}\right),\\
    M_{\mathrm{sup}}(i_x,i_y)
    &=\mathbf{1}\!\left[\mathbf{x}^{\mathrm{2D}}_{i_x,i_y}
      \in\Omega_{\mathrm{sup}}\right].
\end{aligned}
\end{equation}
In~\eqref{eq:support-mask}, \(\mathbf{x}^{\mathrm{2D}}_{i_x,i_y}\) is the center of planar cell \((i_x,i_y)\), \(\alpha_{\mathrm{hull}}\) is the hull parameter, and \(M_{\mathrm{sup}}=1\) denotes a reconstruction-supported cell. Let \(\mathbb{B}(r_{\mathrm{agent}})\) be a disk with the configured agent radius. Combining obstacle inflation with the support mask gives the final navigation costmap
\begin{equation}
    \label{eq:navigation-costmap}
    \mathcal{M}_{\mathrm{occ}}
    =\operatorname{Dilate}\!\left(M_{\mathrm{raw}},\mathbb{B}(r_{\mathrm{agent}})\right)
    \lor \neg M_{\mathrm{sup}}.
\end{equation}
Thus, occupied cells in~\eqref{eq:navigation-costmap} represent obstacles intersecting the configured body, while cells outside \(\Omega_{\mathrm{sup}}\) are treated as unknown and unavailable. Only cells satisfying \(\mathcal{M}_{\mathrm{occ}}=0\) are eligible for start and goal sampling; task-specific connected-reachability and collision checks subsequently determine whether a sampled pair is retained. The resulting costmap supports collision checking, reachability analysis, path planning, and episode filtering.

\FloatBarrier

\subsection{Semantic Grounding for ObjectNav}
\label{sec:semantic}

\paragraph{Problem Setup}
\label{sec:semantic:overview}

The semantic layer converts multi-view open-vocabulary segmentations into ObjectNav instance candidates. The semantic layer prioritizes precision because false-positive centers can create invalid goals, whereas false negatives primarily reduce the candidate pool. This trade-off favors reliable retained episodes over exhaustive object coverage.

\begin{figure}[t]
\centering
\includegraphics[width=\linewidth]{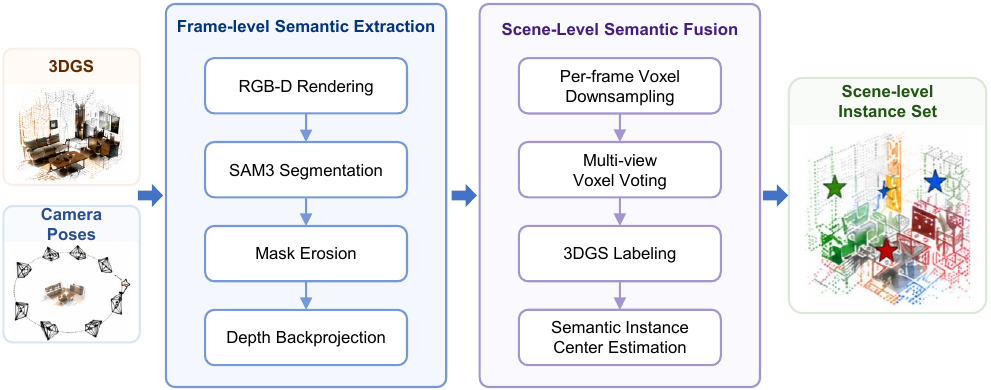}
\caption{
Training-free ObjectNav semantic grounding.
Multi-view open-vocabulary masks are eroded, backprojected, fused by voxel voting, transferred to Gaussian primitives, and clustered into instance centers.
}
\label{fig:semantic-pipeline}
\end{figure}

Given \(\mathcal{G}\), category set \(\mathcal{C}\), and covering poses, NavArena extracts \(\mathcal{I}^{(c)}=\{(\mathbf{x}^{\mathrm{obj}}_{j,c},\ell_{j,c})\}_{j=1}^{J_c}\), where \(\mathbf{x}^{\mathrm{obj}}_{j,c}\) is an instance center and \(\ell_{j,c}=c\). Their union \(\mathcal{I}=\bigcup_{c\in\mathcal{C}}\mathcal{I}^{(c)}\) forms the semantic layer. Figure~\ref{fig:semantic-pipeline} summarizes the grounding process.

\paragraph{Frame-Level Semantic Extraction}
\label{sec:semantic:frame_level}

The first stage associates 2D semantic priors with 3D spatial points. NavArena renders pixel-aligned RGB and depth from the covering poses $\{q_k^{\mathrm{view}}\}$ using differentiable Gaussian rasterization~\cite{ye2025gsplat}.

Rendered views and target-category prompts are passed to SAM3~\cite{carion2025sam3}, which produces per-frame semantic masks. Each mask is eroded before backprojection to suppress noisy boundary pixels caused by opacity bleeding and depth uncertainty. Pixels with valid depth are backprojected through the camera intrinsics and extrinsics, producing a labeled point cloud $\mathcal{P}_k^{(c)}$ for each frame \(k\) and category \(c\).

\paragraph{Scene-Level Semantic Fusion}
\label{sec:semantic:scene_level}

The second stage fuses independent frame observations into scene-level semantic evidence. Each frame cloud is voxel-downsampled at resolution \(\delta_{\mathrm{down}}\), yielding \(\tilde{\mathcal{P}}_k^{(c)}\); their union \(\mathcal{P}^{(c)}=\bigcup_k\tilde{\mathcal{P}}_k^{(c)}\) is evaluated on a coarser voting grid with resolution \(\delta_{\mathrm{vote}}\). Voxels without sufficient support from the pooled multi-view point clouds are discarded; each retained voxel receives the majority label among its points:
\begin{equation}
  \label{eq:semantic-voxel-vote}
  \ell(\mathbf{b}) =
  \operatorname*{arg\,max}_{c \in \mathcal{C}}
  \left|
  \left\{
  \mathbf{p} \in \mathcal{P}^{(c)}
  \mid
  \left\lfloor \mathbf{p} / \delta_{\mathrm{vote}} \right\rfloor = \mathbf{b}
  \right\}
  \right|.
\end{equation}
In~\eqref{eq:semantic-voxel-vote}, \(\mathbf{b}\in\mathbb{Z}^{3}\) denotes a voting-grid voxel index. Voting ties are treated as ambiguous and excluded. The remaining voxel labels are transferred to Gaussian centers according to their enclosing cells. For each category, DBSCAN clusters the labeled centers into object instances, and clusters below the minimum size threshold are discarded. The centroid of each remaining cluster defines $\mathbf{x}^{\mathrm{obj}}_{j,c}$. Aggregating over all categories yields the scene-level semantic candidate set \(\mathcal{I}\).
Each pair $(\mathbf{x}^{\mathrm{obj}}_{j,c}, \ell_{j,c})$ in $\mathcal{I}$ is a semantic candidate rather than a navigation endpoint. The episode generator accepts a candidate only after identifying reachable, collision-valid goal states under the checks in Section~\ref{sec:data:tasks}.

\section{Task Generation and Closed-Loop Evaluation}
\label{sec:generation-evaluation}

\subsection{Task Definition}
\label{sec:data}
\label{sec:data:framework}
\label{sec:data:tasks}
\label{sec:data:format}

An episode is represented as \((q_0,\gamma_{\mathrm{task}},\tau^\star)\) together with an observation specification, where \(q_0\) is the initial robot state, \(\gamma_{\mathrm{task}}\) is the task-specific goal, and \(\tau^\star\) is an expert rollout. The robot state lies in \(\mathrm{SE}(2)\), and the observation specification defines the camera parameters used to render egocentric views from \(\mathcal{G}\).

The three navigation tracks differ only in their goal payloads and success predicates. PointNav specifies a metric target coordinate \(\mathbf{x}_{\mathrm{goal}}=(x_{\mathrm{goal}},y_{\mathrm{goal}})\). ImageNav provides a reference image rendered at a reachable target pose. ObjectNav specifies a target category represented by the semantic candidates defined in Section~\ref{sec:semantic}. This explicit separation preserves task-specific goals while allowing all tracks to share the same scene dynamics and validity checks.

\subsection{Episode Generation}
\label{sec:data:planning}

The generator first samples \(q_0\) from the valid region defined in Section~\ref{sec:scenes:constraints} and then instantiates a task-specific goal. A start--goal pair is accepted only if it satisfies the prescribed distance, reachability, collision, and semantic constraints. For ObjectNav, a semantic candidate is retained only when the generator identifies a reachable, collision-valid goal state associated with the requested object category.

For each accepted pair, NavArena synthesizes a discrete-time expert rollout \(\tau^\star=(\{q_t\}_{t=0}^{T},\{u_t\}_{t=0}^{T-1})\) in \(\mathrm{SE}(2)\) subject to the spatial constraints encoded by \(\mathcal{M}_{\mathrm{occ}}\). The control \(u_t\) is expressed either as a relative waypoint command \((\Delta x,\Delta y,\Delta\theta)\) in the robot-local frame or as a velocity command \((v,\omega,\Delta t)\), where \(\Delta t\) is the integration interval. The generator retains a rollout only if its states and controls satisfy clearance, terminal-goal consistency, heading continuity, and kinematic-feasibility checks. Because observations are rendered on demand from \(\mathcal{G}\), each stored episode comprises the trajectory, task goal, and rendering metadata rather than pre-rendered observation sequences.

\subsection{Closed-loop Evaluation Protocol}
\label{sec:eval}
\label{sec:eval:architecture}
\label{sec:eval:criteria}
\label{sec:eval:metrics}

Closed-loop evaluation executes policy actions in the constructed scenes rather than scoring isolated predictions. Each rollout is initialized with a start state, a task goal, and a step budget. At step \(t\), the simulator renders an egocentric observation from \(\mathcal{G}\), and the policy emits a discrete action \(a_t\in\mathcal{A}_{\mathrm{move}}\cup\{\mathrm{STOP}\}\). The interface layer maps each motion action in \(\mathcal{A}_{\mathrm{move}}\) to a relative waypoint command \((\Delta x,\Delta y,\Delta\theta)\), queries \(\mathcal{M}_{\mathrm{occ}}\), and then updates the robot state. A rejected motion is recorded as a collision or stall. The \(\mathrm{STOP}\) action leaves the robot pose unchanged and is handled according to the active termination protocol.

Within each task track, all policies use the same transition function, action interface, collision model, and metric implementation. The distance-based protocol ignores \(\mathrm{STOP}\) as a termination request: the action consumes one evaluation step without changing the pose, and the rollout succeeds immediately upon entering the success region. Under the explicit-stop protocol, entering the success region alone does not terminate the rollout; success requires \(\mathrm{STOP}\) inside the region, whereas \(\mathrm{STOP}\) outside the region terminates the rollout as a failure. All task-success distances are planar Euclidean distances in the shared metric frame. PointNav measures distance to the target coordinate, ImageNav to the reference-pose position, and ObjectNav to the nearest retained target-instance center projected onto the ground plane. ObjectNav success does not impose an additional target-visibility requirement. Pose SR additionally evaluates heading consistency with the ImageNav goal view, and Stop SR applies the explicit-stop protocol to the ObjectNav success region. Section~\ref{sec:exp:model-eval} reports the corresponding numerical thresholds.

\subsection{Evaluation Metrics}

NavArena reports standard navigation metrics~\cite{anderson2018evaluation} at both aggregate and per-difficulty levels. Distance SR reports success under automatic distance-based termination, whereas Stop SR requires an explicit \(\mathrm{STOP}\) inside the success region. Because the distance-based protocol terminates a rollout immediately upon entering its success region and does not require \(\mathrm{STOP}\), oracle success would be identical to Distance SR and is therefore not reported. NE uses the final planar Euclidean distance, whereas SPL uses the optimal costmap geodesic for path-length normalization. Collision Rate (CR) denotes the fraction of attempted policy actions rejected as collision-invalid, while Average Steps (AvgSteps) measures rollout length. Pose SR reports the stricter ImageNav heading criterion. DTW and NDTW are retained as optional trajectory-shape diagnostics rather than primary benchmark metrics.
\section{Experiments}
\label{sec:exp}

\subsection{Experimental Setup}
\label{sec:exp:setup}

NavArena contains more than 2{,}000 heterogeneous scenes: approximately 1{,}000 artist-created InteriorGS scenes~\cite{sage3d2026}, 1{,}000 mixed-source SceneSplat++ reconstructions~\cite{ma2025scenesplatpp}, and 50 real-world captures collected for this study. Before applying NavArena, the source scenes are manually screened for obvious rendering artifacts and incomplete reconstruction, and scenes failing this check are excluded. This dataset-curation step is external to the automated benchmark-construction pipeline. NavArena is applied only to inputs satisfying the validity and coordinate-frame assumptions stated in Section~\ref{sec:scenes:rep_design}. All sources use the same construction parameters and expose the same rendering, collision, reachability, semantic-goal, and episode-generation interfaces.

We evaluate spatial-query agreement and efficiency, ObjectNav candidate-center precision, dataset scale after validity filtering, and closed-loop policy behavior. These experiments assess benchmark feasibility and diagnostic value rather than complete physical or semantic fidelity.

We evaluate representative policies across PointNav, ImageNav, ObjectNav, and multi-task navigation. The PointNav track includes LoGoPlanner~\cite{peng2025logoplanner}; the ImageNav track includes GNM~\cite{shah2023gnm}, ViNT~\cite{shah2023vint}, NoMaD~\cite{sridhar2024nomad}, and NaviBridger~\cite{ren2025navibridger}; the ObjectNav track includes PoliFormer~\cite{zeng2024poliformer}, Uni-NaVid~\cite{zhang2024uninavid}, and OmniNav~\cite{xue2025omninav}; and the multi-task track includes the PointNav and ImageNav branches of NavDP~\cite{cai2025navdp} and OmniVLA~\cite{hirose2025omnivla}.

Experiments that require additional supervision, such as collision meshes or manual semantic annotations, use only scenes for which the corresponding reference is available.

Within each task track, all policies are evaluated on the same 600 episodes using identical difficulty-dependent step budgets, rendering settings, collision checker, action interface, and metric implementation. Each model retains its native input modality and output representation up to the interface layer. The interface maps non-stop motion actions to the shared waypoint command \((\Delta x,\Delta y,\Delta\theta)\) and preserves \(\mathrm{STOP}\) as a discrete action. Because this adaptation can affect absolute performance, comparisons are restricted to models within the same task track. Results across different tracks are reported only to illustrate distinct failure profiles.

\subsection{Spatial and Semantic Validity}
\label{sec:exp:spatial}
\label{sec:exp:semantic}

We first characterize the spatial layer against a mesh-based reference. NavArena's occupancy-grid queries are compared with FCL queries~\cite{pan2012fcl} on a diversity-sampled subset of 100 scenes with available collision meshes. Each scene contributes 5{,}000 randomly sampled state queries and 1{,}000 motion queries. All methods use the same metric world frame and the same embodiment configuration assigned to each query. In particular, both the occupancy grid and FCL use the body interval \(\mathcal{H}_{\mathrm{body}}\) and footprint \(\mathbb{B}(r_{\mathrm{agent}})\) supplied by that configuration. The embodiment configuration is held fixed across the compared collision backends and policy rollouts. Unless otherwise specified, the occupancy grid uses the default \(0.05\,\mathrm{m}\) resolution.

\begin{table}[t]
  \caption{%
    Spatial-layer query performance at \(0.05\,\mathrm{m}\) grid resolution; lower latency indicates cheaper closed-loop collision and motion-validity checks.
  }
  \label{tab:costmap-main}
  \centering
  \small
  \setlength{\tabcolsep}{2.2pt}
  \begin{tabular}{ccc}
    \toprule
    Query
      & CPU grid ($\mu$s)
      & FCL mesh ($\mu$s) \\
    \midrule
    State  & \textbf{0.986} & 7.802 \\
    Motion & \textbf{39.65} & 425.02 \\
    \bottomrule
  \end{tabular}
\end{table}

\begin{table}[t]
  \caption{%
    Occupancy-grid resolution trade-off against the FCL reference.
    FP and FN denote false-positive and false-negative rates; low FN is preferred because missed collisions create invalid episodes.
  }
  \label{tab:costmap-resolution}
  \centering
  \small
  \setlength{\tabcolsep}{2.7pt}
  \begin{tabular}{cccccc}
    \toprule
    \shortstack{Resolution\\(m)}
      & \shortstack{State FP\\(\%)}
      & \shortstack{State FN\\(\%)}
      & \shortstack{Motion FP\\(\%)}
      & \shortstack{Motion FN\\(\%)}
      & \shortstack{Storage\\(KiB)} \\
    \midrule
    0.02 & \textbf{3.522} & 0.000 & \textbf{2.630} & 0.000 & 382.4 \\
    0.05 & 6.034 & 0.000 & 4.710 & 0.000 & 61.6 \\
    0.10 & 13.054 & 0.002 & 8.590 & 0.000 & \textbf{15.5} \\
    \bottomrule
  \end{tabular}
\end{table}

Tables~\ref{tab:costmap-main} and~\ref{tab:costmap-resolution} show that CPU grid queries are \(7.91\times\) faster for state checks and \(10.72\times\) faster for motion checks than FCL. No motion false negatives occur at the tested resolutions; state false negatives appear only at \(0.10\,\mathrm{m}\) (\(0.002\%\)). The remaining disagreements are predominantly conservative false positives. We adopt \(0.05\,\mathrm{m}\), which retains the observed zero false-negative rate of the finer grid while reducing storage from \(382.4\) to \(61.6~\mathrm{KiB}\). These sampled rates do not guarantee performance on unseen geometries.

For semantic validity, we evaluate ObjectNav goal precision rather than general-purpose semantic segmentation. On a diversity-sampled 10-scene subset, predicted centers are manually checked against the corresponding ground-truth object regions. We report valid-center rate, the fraction of predicted ObjectNav centers that fall inside the target object region. This precision-oriented metric matches the requirements of automatic episode generation: false positives can create invalid goals, whereas false negatives mainly reduce the candidate pool. Because SceneSplat~\cite{li2025scenesplat} and FlashSplat~\cite{shen2024flashsplat} do not directly output object centers, all methods use the same center-extraction rule.

\begin{table}[t]
\centering
\small
\setlength{\tabcolsep}{4pt}
\renewcommand{\arraystretch}{1.05}
\caption{Semantic instance-center validity for ObjectNav goal construction.}
\label{tab:center_quality}
\begin{tabular}{lcc}
\toprule
Method
& \shortstack{Valid Center\\(\%) $\uparrow$}
& \shortstack{Time/Scene\\(min) $\downarrow$} \\
\midrule
SceneSplat (Feedforward)
& 65.95
& \textbf{0.97} \\
FlashSplat (Optimization)
& 57.36
& 135.31 \\
Ours
& \textbf{92.25}
& 6.56 \\
\bottomrule
\end{tabular}
\end{table}

NavArena attains a \(92.25\%\) valid-center rate, exceeding the two baselines by \(26.30\) and \(34.89\) percentage points under the same extraction rule (Table~\ref{tab:center_quality}). It also remains substantially faster than the optimization-based baseline. This experiment evaluates the precision of retained ObjectNav candidates for episode construction; exhaustive object coverage and instance recall are outside the scope of this work.

\subsection{Dataset Scale}
\label{sec:exp:dataset-scale}

We next examine the scalability of the common construction pipeline across heterogeneous scene sources. As summarized in Table~\ref{tab:dataset}, NavArena generates 22.2M expert trajectories for PointNav/ImageNav and ObjectNav across more than 2{,}000 scenes, corresponding to over 160K hours of navigation data. All retained scenes expose the same visual, spatial, semantic, and task-generation interfaces.

\begin{table}[t]
  \caption{%
    Scale of the generated NavArena navigation dataset.
  }
  \label{tab:dataset}
  \centering
  \small
  \setlength{\tabcolsep}{4.5pt}
  \begin{tabular}{lcc}
    \toprule
    Metric
      & \shortstack{PointNav/ImageNav}
      & ObjectNav \\
    \midrule
    Episodes
      & 16.6M
      & 5.6M \\
    Mean Path Length (m)
      & 7.35
      & 3.51 \\
    Avg Steps
      & 567
      & 392 \\
    Total Duration (h)
      & 130K
      & 30K \\
    \bottomrule
  \end{tabular}
\end{table}

\subsection{Closed-loop Diagnostic Utility}
\label{sec:exp:model-eval}

Finally, we test whether NavArena exposes distinct policy behaviors under closed-loop execution. Each task track uses a separate 600-episode zero-shot set sampled across source groups, layouts, appearances, task-relevant goals, and navigation horizons. Every set contains 180 easy, 240 medium, and 180 hard episodes, with step budgets of 500, 1000, and 1500, respectively. All rollouts use the embodiment-specific body interval and footprint supplied by the episode configuration, together with RGB observations rendered at \(640 \times 480\). PointNav success requires a final planar Euclidean distance below \(0.5\,\mathrm{m}\). ImageNav uses a \(1\,\mathrm{m}\) planar Euclidean distance threshold and, when reported, a \(30^{\circ}\) pose criterion. ObjectNav uses a \(1\,\mathrm{m}\) planar Euclidean distance threshold to the nearest retained target-instance center. Termination follows the protocol in Section~\ref{sec:eval}.

\subsubsection{Aggregate Zero-shot Performance}
\label{sec:exp:model-eval:sota}

Table~\ref{tab:model-zs} reports distance-based zero-shot results aggregated over the three difficulty splits. We compare policies only within the same task track; cross-track values illustrate how the evaluation protocol captures different goal-conditioned failure profiles and should not be interpreted as a unified leaderboard.

\begin{table}[t]
  \caption{%
    Aggregate distance-based zero-shot closed-loop performance on NavArena.
  }
  \label{tab:model-zs}
  \centering
  \scriptsize
  \setlength{\tabcolsep}{1.6pt}
  \renewcommand{\arraystretch}{1.04}
  \begin{tabular}{@{}llccccc@{}}
    \toprule
    Task & Model
      & Dist. SR $\uparrow$ & SPL $\uparrow$
      & NE $\downarrow$ & CR $\downarrow$
      & AvgSteps \\
    \midrule
    PointNav
      & LoGoPlanner & 0.679 & 0.597 & 1.601 & 0.470 & 78.2 \\
    \midrule
    \multirow[c]{4}{*}{ImageNav}
      & NaviBridger & \textbf{0.434} & \textbf{0.389} & 2.095 & \textbf{0.694} & \textbf{694.2} \\
      & GNM & \underline{0.403} & 0.362 & \textbf{2.082} & 0.710 & \underline{712.0} \\
      & NoMaD & 0.385 & \underline{0.364} & 2.160 & \underline{0.707} & 734.9 \\
      & ViNT & 0.288 & 0.286 & 2.203 & 0.728 & 806.0 \\
    \midrule
    \multirow[c]{3}{*}{ObjectNav}
      & Uni-NaVid & \textbf{0.374} & \textbf{0.175} & \textbf{2.491} & 0.183 & \underline{163.6} \\
      & PoliFormer & \underline{0.310} & \underline{0.136} & \underline{2.728} & \underline{0.128} & 318.7 \\
      & OmniNav & 0.129 & 0.090 & 3.047 & \textbf{0.108} & \textbf{122.1} \\
    \midrule
    \multirow[c]{4}{*}{Multi-Task}
      & NavDP ImageNav & \textbf{0.661} & \textbf{0.561} & \textbf{1.975} & \textbf{0.545} & \textbf{93.4} \\
      & OmniVLA ImageNav & 0.235 & 0.235 & 2.609 & 0.675 & 125.7 \\
    \cmidrule(l){2-7}
      & NavDP PointNav & \textbf{0.588} & \textbf{0.582} & \textbf{1.429} & \textbf{0.355} & \textbf{87.7} \\
      & OmniVLA PointNav & 0.127 & 0.126 & 2.579 & 0.853 & 138.7 \\
    \bottomrule
  \end{tabular}
\end{table}

Table~\ref{tab:model-zs} reveals distinct within-track profiles. NaviBridger leads single-task ImageNav in Distance SR and CR; Uni-NaVid leads ObjectNav in Distance SR and SPL, although OmniNav has lower CR; and NavDP outperforms OmniVLA on both multi-task branches. AvgSteps is diagnostic rather than an efficiency ranking because early failure may shorten rollouts. LoGoPlanner is a geometry-conditioned PointNav reference and is not directly comparable across goal modalities. Overall, the protocol separates distance-based success, path efficiency, and collision behavior under task-specific interfaces.

\subsubsection{Difficulty-Induced Horizon Degradation}
\label{sec:exp:model-eval:scaling}

The difficulty analysis tests how policy performance changes as closed-loop horizons and action complexity increase while the task interface remains fixed.

\begin{figure}[t]
  \centering
  \includegraphics[width=\columnwidth]{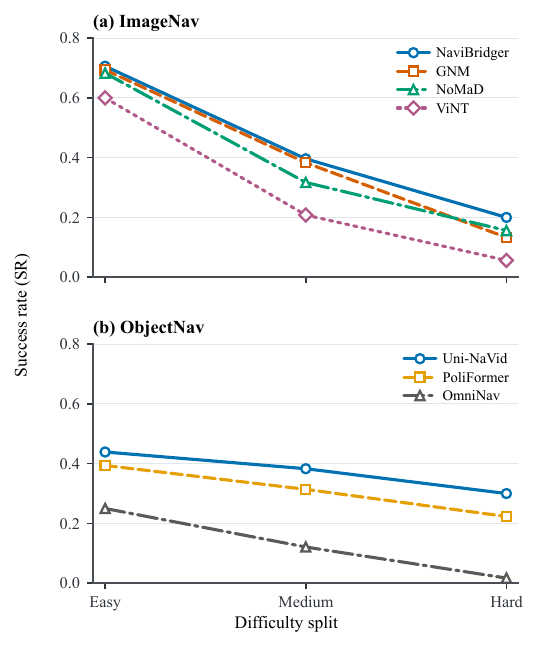}
  \caption{%
    Success degradation across navigation difficulty; longer optimal paths expose horizon-dependent failure modes under the same task interfaces.
  }
  \label{fig:difficulty_scaling}
\end{figure}

Figure~\ref{fig:difficulty_scaling} shows that Distance SR decreases as episode difficulty increases, although the rate of degradation varies substantially across policies. From the easy to the hard split, the single-task ImageNav policies lose \(72\%\)--\(91\%\) of their Distance SR. This trend is compatible with the accumulation of perception and local-control errors over longer horizons, but the present experiment does not isolate their individual contributions. Uni-NaVid and PoliFormer show smaller relative drops of \(32\%\) and \(43\%\), respectively, on the evaluated split, while OmniNav remains weak on hard episodes.

\subsubsection{Termination Sensitivity}
\label{sec:exp:model-eval:ablation}

Distance-based success can conceal last-mile localization and termination failures. We therefore evaluate the same episode sets under the ImageNav pose criterion and rerun ObjectNav with the explicit-stop termination protocol. Table~\ref{tab:termination-diagnostics} reports distance SR, ImageNav pose SR and conditional Pose Retention, and ObjectNav stop-aware SR.

\begin{table}[t]
  \caption{%
    Termination sensitivity under pose and stop criteria. Distance SR succeeds upon entering the target region without requiring \(\mathrm{STOP}\); Stop SR requires \(\mathrm{STOP}\) inside the region.
  }
  \label{tab:termination-diagnostics}
  \centering
  \scriptsize
  \setlength{\tabcolsep}{1.8pt}
  \renewcommand{\arraystretch}{1.04}
  \begin{tabular}{@{}llcccc@{}}
    \toprule
    Track & Model
      & Dist. SR $\uparrow$ & Pose SR $\uparrow$
      & Pose Ret. $\uparrow$ & Stop SR $\uparrow$ \\
    \midrule
    ImageNav & NaviBridger & \textbf{0.434} & \textbf{0.255} & 0.588 & $\mathrm{n/a}$ \\
    ImageNav & GNM         & 0.403 & 0.250 & 0.620 & $\mathrm{n/a}$ \\
    ImageNav & NoMaD       & 0.385 & 0.189 & 0.491 & $\mathrm{n/a}$ \\
    ImageNav & ViNT        & 0.288 & 0.180 & \textbf{0.625} & $\mathrm{n/a}$ \\
    \midrule
    ObjectNav & Uni-NaVid  & \textbf{0.374} & $\mathrm{n/a}$ & $\mathrm{n/a}$ & 0.036 \\
    ObjectNav & PoliFormer & 0.310 & $\mathrm{n/a}$ & $\mathrm{n/a}$ & 0.033 \\
    ObjectNav & OmniNav    & 0.129 & $\mathrm{n/a}$ & $\mathrm{n/a}$ & \textbf{0.048} \\
    \bottomrule
  \end{tabular}
\end{table}

Table~\ref{tab:termination-diagnostics} identifies failures that distance-only SR does not capture. For ImageNav, imposing the heading criterion retains only \(49.1\%\)--\(62.5\%\) of distance-successful episodes, showing that reaching the goal neighborhood does not necessarily imply correct final alignment. Pose Retention is normalized by each model's distance SR and is therefore not biased toward models with lower base success. For ObjectNav, which has no canonical target heading, pose-based metrics are not defined. Stop-aware SR remains below \(0.05\) for all evaluated ObjectNav policies, indicating that successful approach rarely coincides with an explicit stop inside the target neighborhood.

\section{Conclusion and Limitations}
\label{sec:conclusion}

NavArena constructs goal-oriented navigation benchmarks from fixed 3DGS reconstructions by combining a frozen visual layer with derived spatial and semantic representations. The unified construction pipeline supports scalable episode generation, while closed-loop evaluation distinguishes success, collision, horizon, and termination outcomes. Physical and semantic fidelity remains dependent on the underlying representations.
NavArena currently targets mostly static scenes and an \(\mathrm{SE}(2)\) mobile-motion model. Rendering quality is treated as an input condition; the current results may not extend to low-quality reconstructions. Its occupancy and semantic layers remain sensitive to reconstruction and segmentation errors. Policy-specific input and action adaptations may also affect absolute performance. Future work will extend the framework to dynamic environments and richer motion models.

\bibliographystyle{IEEEtran}
\bibliography{ref}

@inproceedings{ramakrishnan2021hm3d,
  title     = {Habitat-{M}atterport {3D} Dataset ({HM3D}): 1000 Large-Scale {3D} Environments for Embodied {AI}},
  author    = {Ramakrishnan, Santhosh Kumar and Gokaslan, Aaron and Wijmans, Erik and Maksymets, Oleksandr and Clegg, Alex and Turner, John and Undersander, Eric and Galuba, Wojciech and Westbury, Andrew and Chang, Angel X and others},
  booktitle = {Advances in Neural Information Processing Systems (NeurIPS)},
  year      = {2021},
}

@inproceedings{deitke2022procthor,
  title     = {{ProcTHOR}: Large-Scale Embodied {AI} Using Procedural Generation},
  author    = {Deitke, Matt and VanderBilt, Eli and Herrasti, Alvaro and Weihs, Luca and Ehsani, Kiana and Salvador, Jordi and Han, Winson and Kolve, Eric and Farhadi, Ali and Kembhavi, Aniruddha and Mottaghi, Roozbeh},
  booktitle = {Advances in Neural Information Processing Systems (NeurIPS)},
  year      = {2022},
}

@article{sage3d2026,
  title   = {Towards Physically Executable {3D} {G}aussian for Embodied Navigation},
  author  = {Miao, Bingchen and Wei, Rong and Ge, Zhiqi and Sun, Xiaoquan and Gao, Shiqi and Zhu, Jingzhe and Wang, Renhan and Tang, Siliang and Xiao, Jun and Tang, Rui and Li, Juncheng},
  journal = {arXiv preprint arXiv:2510.21307},
  year    = {2025},
  doi     = {10.48550/arXiv.2510.21307},
  note    = {InteriorGS: 1{,}000 annotated 3DGS scenes},
}

@article{kerbl2023gaussian,
  title   = {{3D} {G}aussian {S}platting for Real-Time Radiance Field Rendering},
  author  = {Kerbl, Bernhard and Kopanas, Georgios and Leimk{\"u}hler, Thomas and Drettakis, George},
  journal = {ACM Transactions on Graphics (SIGGRAPH)},
  volume  = {42},
  number  = {4},
  year    = {2023},
}

@article{ma2025scenesplatpp,
  title   = {{SceneSplat}++: A Large Dataset and Comprehensive Benchmark for Language {G}aussian Splatting},
  author  = {Ma, Mengjiao and Ma, Qi and Li, Yue and Cheng, Jiahuan and Yang, Runyi and Ren, Bin and Popovic, Nikola and Wei, Mingqiang and Sebe, Nicu and Van Gool, Luc and Gevers, Theo and Oswald, Martin R. and Paudel, Danda Pani},
  journal = {arXiv preprint arXiv:2506.08710},
  year    = {2025},
  doi     = {10.48550/arXiv.2506.08710},
  note    = {SceneSplat-49K: approx.\ 49K raw / 46K curated 3DGS scenes, indoor and outdoor},
}

@inproceedings{qin2024langsplat,
  title     = {{LangSplat}: {3D} Language {G}aussian Splatting},
  author    = {Qin, Minghan and Li, Wanhua and Zhou, Jiawei and Wang, Haoqian and Pfister, Hanspeter},
  booktitle = {IEEE/CVF Conference on Computer Vision and Pattern Recognition (CVPR)},
  year      = {2024},
}

@inproceedings{shah2023vint,
  title     = {{ViNT}: A Foundation Model for Visual Navigation},
  author    = {Shah, Dhruv and Sridhar, Ajay and Dashora, Nitish and Stachowicz, Kyle and Black, Kevin and Hirose, Noriaki and Levine, Sergey},
  booktitle = {Conference on Robot Learning (CoRL)},
  year      = {2023},
}

@inproceedings{sridhar2024nomad,
  title     = {{NoMaD}: Goal Masked Diffusion Policies for Navigation and Exploration},
  author    = {Sridhar, Ajay and Shah, Dhruv and Glossop, Catherine and Levine, Sergey},
  booktitle = {IEEE International Conference on Robotics and Automation (ICRA)},
  year      = {2024},
}

@inproceedings{shah2023gnm,
  title     = {{GNM}: A General Navigation Model to Drive Any Robot},
  author    = {Shah, Dhruv and Sridhar, Ajay and Bhorkar, Arjun and Hirose, Noriaki and Levine, Sergey},
  booktitle = {IEEE International Conference on Robotics and Automation (ICRA)},
  year      = {2023},
}

@article{zeng2024poliformer,
  title   = {{PoliFormer}: Scaling On-Policy {RL} with Transformers Results in Masterful Navigators},
  author  = {Zeng, Kuo-Hao and Zhang, Zichen and Ehsani, Kiana and Hendrix, Rose and Salvador, Jordi and Herrasti, Alvaro and Girshick, Ross and Kembhavi, Aniruddha and Weihs, Luca},
  journal = {arXiv preprint arXiv:2406.20083},
  year    = {2024},
  doi     = {10.48550/arXiv.2406.20083},
}

@article{zhang2024uninavid,
  title   = {{Uni-NaVid}: A Video-Based Vision-Language-Action Model for Unifying Embodied Navigation Tasks},
  author  = {Zhang, Jiazhao and Wang, Kunyu and Wang, Shaoan and Li, Minghan and Liu, Haoran and Wei, Songlin and Wang, Zhongyuan and Zhang, Zhizheng and Wang, He},
  journal = {arXiv preprint arXiv:2412.06224},
  year    = {2024},
  doi     = {10.48550/arXiv.2412.06224},
}

@article{xue2025omninav,
  title   = {{OmniNav}: A Unified Framework for Prospective Exploration and Visual-Language Navigation},
  author  = {Xue, Xinda and Hu, Junjun and Luo, Minghua and Xie, Shichao and Chen, Jintao and Xie, Zixun and Quan, Kuichen and Guo, Wei and Xu, Mu and Chu, Zedong},
  journal = {arXiv preprint arXiv:2509.25687},
  year    = {2025},
  doi     = {10.48550/arXiv.2509.25687},
}

@article{ren2025navibridger,
  title   = {Prior Does Matter: Visual Navigation via Denoising Diffusion Bridge Models},
  author  = {Ren, Hao and Zeng, Yiming and Bi, Zetong and Wan, Zhaoliang and Huang, Junlong and Cheng, Hui},
  journal = {arXiv preprint arXiv:2504.10041},
  year    = {2025},
  doi     = {10.48550/arXiv.2504.10041},
}

@article{cai2025navdp,
  title   = {{NavDP}: Learning Sim-to-Real Navigation Diffusion Policy with Privileged Information Guidance},
  author  = {Cai, Wenzhe and Peng, Jiaqi and Yang, Yuqiang and Zhang, Yujian and Wei, Meng and Wang, Hanqing and Chen, Yilun and Wang, Tai and Pang, Jiangmiao},
  journal = {arXiv preprint arXiv:2505.08712},
  year    = {2025},
  doi     = {10.48550/arXiv.2505.08712},
}

@inproceedings{hirose2025omnivla,
  title     = {{OmniVLA}: An Omni-Modal Vision-Language-Action Model for Robot Navigation},
  author    = {Hirose, Noriaki and Glossop, Catherine and Shah, Dhruv and Levine, Sergey},
  booktitle = {IEEE International Conference on Robotics and Automation (ICRA)},
  year      = {2026},
}

@article{peng2025logoplanner,
  title   = {{LoGoPlanner}: Localization Grounded Navigation Policy with Metric-Aware Visual Geometry},
  author  = {Peng, Jiaqi and Cai, Wenzhe and Yang, Yuqiang and Wang, Tai and Shen, Yuan and Pang, Jiangmiao},
  journal = {arXiv preprint arXiv:2512.19629},
  year    = {2025},
  doi     = {10.48550/arXiv.2512.19629},
}

@article{lei2025gaussnav,
  title   = {{GaussNav}: {G}aussian Splatting for Visual Navigation},
  author  = {Lei, Xiaohan and Wang, Min and Zhou, Wengang and Li, Houqiang},
  journal = {IEEE Transactions on Pattern Analysis and Machine Intelligence},
  volume  = {47},
  year    = {2025},
  doi     = {10.1109/TPAMI.2025.3538496},
}

@article{jia2025discoverse,
  title   = {{DISCOVERSE}: Efficient Robot Simulation in Complex High-Fidelity Environments},
  author  = {Jia, Yufei and Wang, Guangyu and Dong, Yuhang and Wu, Junzhe and Zeng, Yupei and Lin, Haonan and Wang, Zifan and Ge, Haizhou and Gu, Weibin and Ding, Kairui and Yan, Zike and Cheng, Yunjie and Li, Yue and Wang, Ziming and Li, Chuxuan and Sui, Wei and Shi, Lu and Tian, Guanzhong and Huang, Ruqi and Zhou, Guyue},
  journal = {arXiv preprint arXiv:2507.21981},
  year    = {2025},
  doi     = {10.48550/arXiv.2507.21981},
  note    = {3DGS + MuJoCo Real2Sim2Real simulator},
}

@article{escontrela2025gaussgym,
  title   = {{GaussGym}: An Open-Source Real-to-Sim Framework for Learning Locomotion from Pixels},
  author  = {Escontrela, Alejandro and Kerr, Justin and Allshire, Arthur and Frey, Jonas and Duan, Rocky and Sferrazza, Carmelo and Abbeel, Pieter},
  journal = {arXiv preprint arXiv:2510.15352},
  year    = {2025},
  doi     = {10.48550/arXiv.2510.15352},
  note    = {3DGS as a drop-in renderer inside IsaacGym for pixel-based locomotion and navigation},
}

@inproceedings{savva2019habitat,
  title     = {Habitat: A Platform for Embodied {AI} Research},
  author    = {Savva, Manolis and Kadian, Abhishek and Maksymets, Oleksandr and Zhao, Yili and Wijmans, Erik and Jain, Bhavana and Straub, Julian and Liu, Jia and Koltun, Vladlen and Malik, Jitendra and Batra, Dhruv and Parikh, Devi},
  booktitle = {IEEE/CVF International Conference on Computer Vision (ICCV)},
  year      = {2019},
}

@article{carion2025sam3,
  title   = {{SAM} 3: Segment Anything with Concepts},
  author  = {Carion, Nicolas and Gustafson, Laura and Hu, Yuan-Ting and Debnath, Shoubhik and Hu, Ronghang and Suris, Didac and Ryali, Chaitanya and Alwala, Kalyan Vasudev and Khedr, Haitham and Huang, Andrew and Lei, Jie and Ma, Tengyu and Guo, Baishan and Kalla, Arpit and Marks, Markus and Greer, Joseph and Wang, Meng and Sun, Peize and R{\"a}dle, Roman and Afouras, Triantafyllos and others},
  journal = {arXiv preprint arXiv:2511.16719},
  year    = {2025},
  doi     = {10.48550/arXiv.2511.16719},
}

@article{anderson2018evaluation,
  title   = {On Evaluation of Embodied Navigation Agents},
  author  = {Anderson, Peter and Chang, Angel and Chaplot, Devendra Singh and Dosovitskiy, Alexey and Gupta, Saurabh and Koltun, Vladlen and Kosecka, Jana and Malik, Jitendra and Mottaghi, Roozbeh and Savva, Manolis and Zamir, Amir R.},
  journal = {arXiv preprint arXiv:1807.06757},
  year    = {2018},
  doi     = {10.48550/arXiv.1807.06757},
}

@misc{Jia2026_EHVJRIBC,
  title         = {{GS-Playground}: A High-Throughput Photorealistic Simulator for Vision-Informed Robot Learning},
  author        = {Jia, Yufei and Zhang, Heng and Zhang, Ziheng and Wu, Junzhe and Yu, Mingrui and Wang, Zifan and Jiang, Dixuan and Li, Zheng and Cao, Chenyu and Yu, Zhuoyuan and Yang, Xun and Ge, Haizhou and Zhang, Yuchi and Zhang, Jiayuan and Huang, Zhenbiao and Liu, Tianle and Chen, Shenyu and Wang, Jiacheng and Xie, Bin and Yao, Xuran and Deng, Xiwa and Wang, Guangyu and Zhang, Jinzhi and Hao, Lei and Chen, Zhixing and Chen, Yuxiang and Wang, Anqi and Tian, Hongyun and Yan, Yiyi and Cao, Zhanxiang and Jiang, Yizhou and Shao, Hanyang and Li, Yue and Shi, Lu and Chen, Bokui and Sui, Wei and Cui, Hanqing and Qin, Yusen and Huang, Ruqi and Han, Lei and Wang, Tiancai and Zhou, Guyue},
  year          = {2026},
  eprint        = {2604.25459},
  archivePrefix = {arXiv},
  primaryClass  = {cs.RO},
  doi           = {10.48550/arXiv.2604.25459},
}

@inproceedings{guedon2024sugar,
  title     = {{SuGaR}: Surface-Aligned {G}aussian Splatting for Efficient {3D} Mesh Reconstruction and High-Quality Mesh Rendering},
  author    = {Gu{\'e}don, Antoine and Lepetit, Vincent},
  booktitle = {IEEE/CVF Conference on Computer Vision and Pattern Recognition (CVPR)},
  year      = {2024}
}

@inproceedings{kerr2023lerf,
  title     = {{LERF}: Language Embedded Radiance Fields},
  author    = {Kerr, Justin and Kim, Chung Min and Goldberg, Ken and Kanazawa, Angjoo and Tancik, Matthew},
  booktitle = {IEEE/CVF International Conference on Computer Vision (ICCV)},
  year      = {2023}
}

@misc{xia2026habitatgs,
  title         = {{Habitat-GS}: A High-Fidelity Navigation Simulator with Dynamic Gaussian Splatting},
  author        = {Xia, Ziyuan and Xu, Jingyi and Cui, Chong and Yu, Yuanhong and Zhang, Jiazhao and Yan, Qingsong and Ni, Tao and Chen, Junbo and Zhou, Xiaowei and Bao, Hujun and Hu, Ruizhen and Peng, Sida},
  year          = {2026},
  eprint        = {2604.12626},
  archivePrefix = {arXiv},
  primaryClass  = {cs.RO},
  doi           = {10.48550/arXiv.2604.12626},
}

@article{batra2020objectnav,
  title   = {{ObjectNav} Revisited: On Evaluation of Embodied Agents Navigating to Objects},
  author  = {Batra, Dhruv and Gokaslan, Aaron and Kembhavi, Aniruddha and Maksymets, Oleksandr and Mottaghi, Roozbeh and Savva, Manolis and Toshev, Alexander and Wijmans, Erik},
  journal = {arXiv preprint arXiv:2006.13171},
  year    = {2020},
  doi     = {10.48550/arXiv.2006.13171},
}

@inproceedings{makoviychuk2021isaac,
  title={Isaac Gym: High Performance {GPU}-Based Physics Simulation For Robot Learning},
  author={Makoviychuk, Viktor and Wawrzyniak, Lukasz and Guo, Yunrong and Lu, Michelle and Storey, Kier and Macklin, Miles and Hoeller, David and Rudin, Nikita and Allshire, Arthur and Handa, Ankur and others},
  booktitle={Thirty-fifth Conference on Neural Information Processing Systems Datasets and Benchmarks Track},
  year={2021}
}

@inproceedings{li2025scenesplat,
  title={Scenesplat: Gaussian splatting-based scene understanding with vision-language pretraining},
  author={Li, Yue and Ma, Qi and Yang, Runyi and Li, Huapeng and Ma, Mengjiao and Ren, Bin and Popovic, Nikola and Sebe, Nicu and Konukoglu, Ender and Gevers, Theo and others},
  booktitle={Proceedings of the IEEE/CVF International Conference on Computer Vision},
  pages={4961--4972},
  year={2025}
}

@inproceedings{li2025instancegaussian,
  title={Instancegaussian: Appearance-semantic joint gaussian representation for 3d instance-level perception},
  author={Li, Haijie and Wu, Yanmin and Meng, Jiarui and Gao, Qiankun and Zhang, Zhiyao and Wang, Ronggang and Zhang, Jian},
  booktitle={Proceedings of the Computer Vision and Pattern Recognition Conference},
  pages={14078--14088},
  year={2025}
}

@inproceedings{shen2024flashsplat,
  title={Flashsplat: 2d to 3d gaussian splatting segmentation solved optimally},
  author={Shen, Qiuhong and Yang, Xingyi and Wang, Xinchao},
  booktitle={European Conference on Computer Vision},
  pages={456--472},
  year={2024},
  organization={Springer}
}

@article{kolve2017ai2thor,
  title   = {{AI2-THOR}: An Interactive {3D} Environment for Visual {AI}},
  author  = {Kolve, Eric and Mottaghi, Roozbeh and Han, Winson and VanderBilt, Eli and Weihs, Luca and Herrasti, Alvaro and Deitke, Matt and Ehsani, Kiana and Gordon, Daniel and Zhu, Yuke and Gupta, Abhinav and Farhadi, Ali},
  journal = {arXiv preprint arXiv:1712.05474},
  year    = {2017},
  doi     = {10.48550/arXiv.1712.05474},
}

@inproceedings{xia2018gibson,
  title     = {Gibson Env: Real-World Perception for Embodied Agents},
  author    = {Xia, Fei and Zamir, Amir R. and He, Zhiyang and Sax, Alexander and Malik, Jitendra and Savarese, Silvio},
  booktitle = {IEEE/CVF Conference on Computer Vision and Pattern Recognition (CVPR)},
  year      = {2018}
}

@article{ye2025gsplat,
  title={gsplat: An open-source library for Gaussian splatting},
  author={Ye, Vickie and Li, Ruilong and Kerr, Justin and Turkulainen, Matias and Yi, Brent and Pan, Zhuoyang and Seiskari, Otto and Ye, Jianbo and Hu, Jeffrey and Tancik, Matthew and others},
  journal={Journal of Machine Learning Research},
  volume={26},
  number={34},
  pages={1--17},
  year={2025}
}

@inproceedings{pan2012fcl,
  title     = {{FCL}: A General Purpose Library for Collision and Proximity Queries},
  author    = {Pan, Jia and Chitta, Sachin and Manocha, Dinesh},
  booktitle = {IEEE International Conference on Robotics and Automation (ICRA)},
  year      = {2012},
}

\end{document}